# BAN-ABSA: An Aspect-Based Sentiment Analysis dataset for Bengali and it's baseline evaluation

Mahfuz Ahmed Masum[1*][0000−0001−6505−7321], Sheikh Junayed Ahmed[1][0000−0002−4429−6726], Ayesha Tasnim[1,2][0000−0002−7143−3255], and Md Saiful Islam[1,2][0000−0001−9236−380X]

[1] Shahjalal University of Science and Technology
Kumargaon, Sylhet 3114, Bangladesh
{mahfuz27, sheikh44} @student.sust.edu
[2] {tasnim-cse,saiful-cse} @sust.edu

**Abstract.** Due to the breathtaking growth of social media or newspaper user comments, online product reviews comments, sentiment analysis (SA) has captured substantial interest from the researchers. With the fast increase of domain, SA work aims not only to predict the sentiment of a sentence or document but also to give the necessary detail on different aspects of the sentence or document (i.e. aspect-based sentiment analysis). A considerable number of datasets for SA and aspect-based sentiment analysis (ABSA) have been made available for English and other well-known European languages. In this paper, we present a manually annotated Bengali dataset of high quality, BAN-ABSA, which is annotated with aspect and its associated sentiment by 3 native Bengali speakers. The dataset consists of 2,619 positive, 4,721 negative and 1,669 neutral data samples from 9,009 unique comments gathered from some famous Bengali news portals. In addition, we conducted a baseline evaluation with a focus on deep learning model, achieved an accuracy of 78.75% for aspect term extraction and accuracy of 71.08% for sentiment classification. Experiments on the BAN-ABSA dataset show that the CNN model is better in terms of accuracy though Bi-LSTM significantly outperforms CNN model in terms of average F1-score.

## 1 Introduction

Newspapers provide a fantastic source of knowledge of different domains. With the extensive growth of the web platform, people express their opinions on various topics, complain and convey sentiment on different social websites or news portals. In this paper, we are going deep down to classify people's subjective opinions based on Bengali newspaper comments. Sentiment analysis [16] tries to extract the biased information from a user-written textual content by classifying it into one of the predefined sets of classes. Aspect-based sentiment analysis (ABSA) is a text classification system that aims to detect the aspect terms in sentences and predict the associated sentiment polarities. For example, the following sentence is taken from a Bengali news portal's comment section:

"Ronaldo and Messi are very good footballers."

This sentence has the aspect of sports and the polarity is positive. Aspects and sentiments may or may not be explicitly mentioned in a sentence. It is the engagement of a machine learning algorithm, to identify the implicitly stated aspect term and sentiment polarity.

The ABSA task can be divided into two sub-tasks: aspect term extraction and sentiment classification. The principal responsibility of opinion mining is aspect term finding [13]. Identification of the terms of aspect is often referred to as aspect term or opinion target extraction. We have used supervised learning-based approaches in this paper for both aspect term extraction and classification of sentiments. The supervised approach to learning is one that establishes a relationship between inputs and outputs, based on examples.

The Bengali language has suffered from a lack of research on the study of ABSA. We devoted our work to the Bengali language by contributing as follows:

- We created a benchmark dataset, BAN-ABSA, that contains 9009 unique comments collected from some popular Bengali news portals. We classified a sentence into one of four aspects: Politics, Sports, Religion, and Others. Again for every aspect, we use respective ternary sentiment classification: Positive, Negative, and Neutral.
- We made the dataset publicly available [3].
- We also conducted a baseline evaluation on our dataset by implementing some state-of-the-art models to find the the one that gives best performance in terms of Bengali ABSA.

## 2  Related Work

We list the most remarkable researches pertaining to both SA and ABSA throughout this section. Due to the deficiency of researches regarding this topic in Bengali, we also mention some works conducted in English.

Sentiment Analysis can be considered as a problem of classification in NLP. Methodologies based on lexicon were used primarily in earlier researches [15, 23]. They involved creating customized features that were expensive. The feature quality significantly influenced the outcome.

ABSA is a fine-grained version of SA that seeks to detect not only the polarity but also the aspect. Key tasks of ABSA include aspect extraction and polarity detection [8]. Deep neural networks are used recently in ABSA with great success. Long Short Term Memory (LSTM) [11], which is a variation of Recurrent Neural Network (RNN) with feedback connections, is being used in many models with promising results [14, 29, 30]. Convolutional Neural Network (CNN) is a neural network that can assign weights and biases to various aspects in the input data using convolutional layers. CNN based models are found to be a good performer in this task [12, 31]. Clause level classification of a sentence can also show significant performance [28]. A language model can distinguish between phrases in a sentence by using probability distribution over the sequence of words. The use of language models have introduced a new dimension to the sentiment analysis [9, 17, 20, 26].

---

[3] https://www.kaggle.com/mahfuzahmed/banabsa

Several datasets have been proposed for ABSA in English and other languages. A dataset was created in [24], namely SentiHood, which contains 5215 sentences collected from Yahoo! question answering system of urban neighborhoods. The dataset has 11 aspects and 3 sentiments. They conducted a baseline evaluation on the dataset and got best accuracy using Logistic Regression (LR) based model LR-Mask and LSTM in the sub-tasks of ABSA respectively. [7] proposed a dataset containing 5412 Hindi product review sentences of 12 domains and 4 polarities. Conditional Random Field (CRF) and Support Vector Machine (SVM) performed the best in their dataset. 39 datasets containing 7 domains of 8 languages were introduced in SemEval-2016 Task 5 [18]. The available languages were English, Arabic, Chinese, Dutch, French, Russian, Spanish, and Turkish. Another ABSA dataset containing 2200 reviews of the Czech language was crafted in [27]. They achieved best performance by utilizing linear-chain CRF.

ABSA is relatively new in Bengali language. Two datasets were introduced in [21], a cricket and a restaurant dataset. The cricket dataset contains 2900 comments collected from online news portals and the restaurant dataset contains 2800 comments translated from an English corpus. The research was taken further in [22] by utilizing CNN to extract the aspect from a sentence. Our research is the first one in Bengali which completes both sub-tasks, aspect extraction, and sentiment classification using BiLSTM neural network.

## 3  BAN-ABSA Description

There is always a scarcity of the Bengali benchmark dataset for Aspect Based Sentiment Analysis. Though two datasets were presented in [21], there were very few comments in both of the datasets. So we created another benchmark dataset for this task. Our BAN-ABSA dataset contains a total of 9009 comments collected from some prominent Bengali online news portals. It contains four aspects: politics, sports, religion, and others and every aspect have their associated polarity: positive, negative, and neutral. The comments are collected by the authors of this paper and annotated in collaboration with 9 annotators.

### 3.1  Data Collection

Our target was to collect comments from various news portals where enormous people express their opinions on different topics. Among several online portals, we opted for the most popular ones. The data collection process is described below:

**Daily Prothom Alo [1]:** Prothom Alo is a very popular Bangladeshi newspaper. Authentic news about daily affairs are published here. On their website, there is a comment section where people post their opinions about ongoing events of all aspects. But very few comments can be found on their website. We head over to their Facebook official page [2] and collected 16735 comments of different aspects.

**Daily Jugantor [3]:** Daily Jugantor is another popular Bangladeshi newspaper but due to the unavailability of comments in their website, we reached out to their Facebook page [4] and collected 14389 comments.

**Kaler Kantho [5]:** Lots of people in Bangladesh read Kaler Kantho online. Very few of them throw their opinions on the website. A total of 8062 comments were collected from the website and Facebook official page [6].

After the data collection process, there were a total of 39186 comments to pre-process.

### 3.2   Annotating Data

After the collection of comments, we removed multi-lined comments from the dataset. In some cases, we trimmed multi-lined comment into single-line comment. We removed emojis from the comments. These were done manually by the authors. After that, 9009 comments remained for annotating.

For the purpose of annotating, we split the data into three parts and divided between 9 annotators. All of them are undergraduate students. Each comment was annotated by 3 annotators for both aspect and sentiment polarity. Any comment could have one of 3 aspects: politics, religion, and sports. The comment that doesn't fall into any of the aforementioned aspects, falls in the other aspect. Again the comment can be either positive, negative, or neutral. If there was any contradiction for any comment, we followed the majority voting. The following comment can be considered as an example:

<div align="center">"এই খুনের পিছনে ধর্ম কোন কাজ করেনি, কাজ করেছে রাজনীতি"</div>

The outcome of annotating this comment by 3 annotators is shown in Table 1.

'

**Table 1.** Annotation example of a comment

| Comment: এই খুনের পিছনে ধর্ম কোন কাজ করেনি, কাজ করেছে রাজনীতি | | |
|---|---|---|
| Annotator | Aspect | Polarity |
| A1 | Politics | Negative |
| A2 | Religion | Negative |
| A3 | Politics | Negative |

We can see from Table 1 that the comment was marked as negative by all the annotators. For the aspect annotation, it was marked as containing Politics aspect by 2 of the annotators and one annotator marked it as containing Religion aspect. From majority voting, we finally agreed to annotate this comment as having **Politics** aspect and **Negative** polarity. We didn't face any tie situation. Table 2 shows a sample of BAN-ABSA.

**Table 2.** Dataset sample

| Comment | Aspect | Polarity |
| --- | --- | --- |
| মেসি চলে যাবে সিটি তে | sports | neutral |
| বার্সেলোনা ওয়ান্ডকাপ জিতবে এইবার | sports | positive |
| ধন্যবাদ, সাধুবাদ ও শুভ কামনা জানাই প্রধান নির্বাচন কমিশনার সাহেব কে | politics | positive |
| এখন মদিনা সনদ আর পাঁচ ওয়াক্ত তাহাজ্জত নামাজ দিয়ে রাষ্ট্র পরিচালিত হবে | religion | neutral |
| বড় আশাছিলো পাক ভারত একটা যুদ্ধ দেখব, তা আর দেখা হলোনা | politics | negative |
| আজ প্রমাণ হলো বাংলাদেশে কোন সংবিধান নেই। | politics | negative |
| আল আমিনের বোলিং তো অসাধারণ | sports | positive |
| মুসলিমের আলোর পথে আনার জন্য তিনি হলেন এক মাত্র ব্যক্তি বর্তমানে। | religion | positive |
| গ্রামীণফোনের ৫০ লাখ ফোরজি গ্রাহক | others | neutral |
| জোর করে রাজ্য জয় করলেন কিন্তু মনটা জয় করতে পারলেন না। | others | negative |

In the final dataset, we have 9009 labeled comments. The dataset contains four aspects and 3 polarities. We tried to balance the dataset as much as possible but in the case of news, people usually throw fewer comments for a piece of good news. Conversely, plenty of comments can be found on a post about any bad incident. The statistics of the dataset are given in Table 3.

'
**Table 3.** Statistics of the dataset

| Aspect | Positive | Negative | Neutral | Subtotal |
| --- | --- | --- | --- | --- |
| Politics | 506 | 1684 | 473 | 2663 |
| Sports | 646 | 407 | 527 | 1580 |
| Religion | 958 | 594 | 201 | 1753 |
| Others | 509 | 2036 | 468 | 3013 |
| Total | | | | 9009 |

### 3.3   Dataset Analysis

In order to understand the annotators' reliability, we calculated Intra-class Correlation Coefficient (ICC) which was 0.77. We also applied Zipf's law [19] in our dataset. Zipf's law is an empirical law proposed by George Kingsley Zipf. It states that, the collection frequency $cf_i$ of the $i$th most common term in the dataset should be proportional to $1/i$:

$$cf_i \propto \frac{1}{i}$$

The application of Zipf's law is shown graphically in figure 1. It indicates that the words are not random in our dataset.

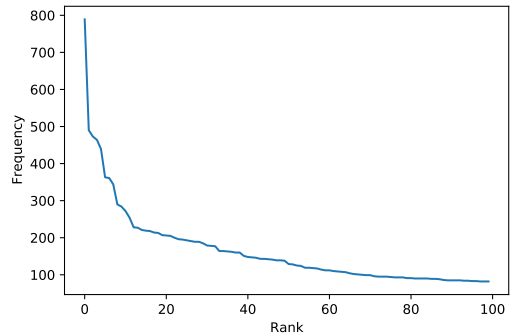

**Fig. 1.** Word frequency distribution

## 4   Experiment

We created a benchmark dataset, BAN-ABSA, to be used in Bengali ABSA. It can be used for both the sub-tasks: aspect extraction and sentiment classification. In the pre-processing step of our experiment, we removed the comments that contained English words. We removed all the punctuation marks. Finally, we tokenized the data. We evaluated some state-of-the-art deep neural network models and some traditional supervised machine learning models to show the baseline for this task. Among several models, we achieved the highest result using Bi-LSTM in both sub-tasks.

### 4.1   Bi-Directional Long Short Term Memory

Recurrent Neural Networks (RNN) are being used in text classification tasks with significant results. To overcome the vanishing gradient problem associated with RNN, a new variant was introduced, LSTM [11]. As for LSTM, information can only be propagated in forward states. As a result, the state is dependent only on past information at any time. But we may need to use forward information as well. BiLSTM has been introduced to overcome that shortcoming. BiLSTM's architecture incorporates two hidden, opposite direction LSTM layers to completely capture the input context. This is achieved through the splitting of a regular RNN's state neurons into two parts. One is used to capture features of the past using forward states and the other one captures features of the opposite direction using backward states [25]. BiLSTM cells are actually unidirectional LSTM cells, without the reverse states. A simple representation of this model is shown in Figure 2.

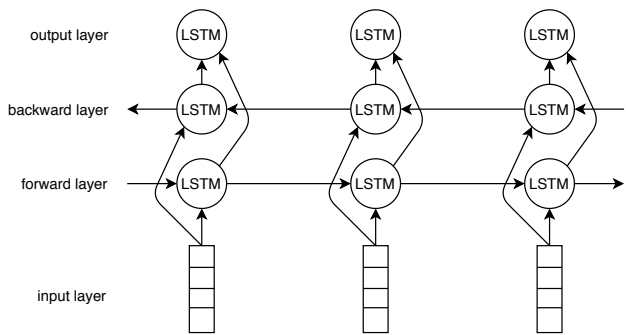

**Fig. 2.** BiLSTM Architecture.

### 4.2 Result and Discussion

The sub-tasks of ABSA are both multi-label classification problem. Deep neural networks work better than conventional supervised machine learning models in such tasks. The results of our evaluation on the dataset for aspect extraction and sentiment classification is shown in Table 4 and 5.

**Table 4.** Performance of aspect extraction on the proposed dataset

| Model | Accuracy | F-1 Score | Precision | Recall |
|---|---|---|---|---|
| SVM | 68.53 | 69.51 | 73.87 | 65.64 |
| CNN | **79.09** | 74.69 | 74.61 | 74.78 |
| LSTM | 77.17 | 77.24 | 78.11 | 76.38 |
| Bi-LSTM | 78.75 | **79.38** | **80.46** | **78.33** |

It is clear that BiLSTM works better than other deep neural networks in Bengali ABSA. It shows significant F-1 and recall score in both aspect extraction

88       M. A. Masum et al.

**Table 5.** Performance of sentiment classification on the proposed dataset

| Model   | Accuracy  | F-1 Score | Precision | Recall    |
|---------|-----------|-----------|-----------|-----------|
| SVM     | 65.02     | 41.15     | 45.13     | 37.82     |
| CNN     | **71.48** | 60.15     | 61.89     | 58.51     |
| LSTM    | 69.51     | 61.20     | **63.37** | 59.17     |
| Bi-LSTM | 71.08     | **62.30** | 62.49     | **62.11** |

and sentiment classification. Though LSTM works very closely with Bi-LSTM, it still fails to catch some aspects. BiLSTM has better context catching capability because of its bi-directional approach. CNN shows better accuracy in both of the sub-tasks, by achieving accuracy score of 79.09% and 71.48% respectively. Bi-LSTM outperforms CNN by achieving a higher f-1 score of 79.38% and 62.30% respectively in the sub-tasks.

To the best of our knowledge, only aspect extraction has been done in Bengali ABSA in [10, 21]. Performance comparison of aspect extraction between Bi-LSTM and other models that were used previously is shown in Table 6. The models were evaluated on our dataset.

**Table 6.** Aspect extraction performance comparison

| Model       | Accuracy  | F-1 Score | Precision | Recall    |
|-------------|-----------|-----------|-----------|-----------|
| SVM [10,21] | 68.53     | 69.51     | 73.87     | 65.64     |
| KNN [10,21] | 50.69     | 47.05     | 58.30     | 45.64     |
| RF [10,21]  | 65.25     | 65.10     | 68.31     | 63.55     |
| CNN [22]    | 79.09     | 74.69     | 74.61     | 74.78     |
| Bi-LSTM     | **78.75** | **79.38** | **80.46** | **78.33** |

**Table 7.** Performance comparison of dataset proposed in [21] and our dataset

| Dataset          | Model | F-1 Score | Precision | Recall   |
|------------------|-------|-----------|-----------|----------|
| Cricket [21]     | SVM   | 0.34      | 0.71      | 0.22     |
|                  | KNN   | 0.25      | 0.45      | 0.21     |
|                  | RF    | 0.37      | 0.60      | 0.27     |
|                  | CNN   | 0.51      | 0.54      | 0.48     |
| Restaurant [21]  | SVM   | 0.38      | **0.77**  | 0.30     |
|                  | KNN   | 0.42      | 0.54      | 0.34     |
|                  | RF    | 0.33      | 0.64      | 0.26     |
|                  | CNN   | 0.64      | 0.67      | 0.61     |
| Proposed dataset | SVM   | **0.69**  | 0.74      | **0.66** |
|                  | KNN   | **0.47**  | **0.58**  | **0.46** |
|                  | RF    | **0.65**  | **0.68**  | **0.64** |
|                  | CNN   | **0.75**  | **0.75**  | **0.75** |

SVM, KNN, RF, and CNN models were evaluated for aspect extraction on the dataset proposed in [21] and our dataset. Table 7 shows the comparison of the performances between our dataset and the dataset stated in [21].

It is visible from Table 7 that quality of dataset matters when it comes to performance. The same models perform better in our dataset as our dataset is bigger in size compared to the Cricket and Restaurant dataset and contains more variation.

Some hyperparameters that we have used to evaluate the models for aspect extraction are given in Table 8.

Table 8. Hyper-parameters used in aspect extraction task

| Model | Hyper-parameters |
|---|---|
| SVM | kernel = rbf <br> regularization parameter = 1.0 |
| KNN | neighbours = 7 <br> weights: uniform <br> distance metric: minkowski |
| RF | number of trees in a forest = 100 <br> all nodes were expanded until all leaves contain less than 2 samples <br> weighted impurity decrease : <br> $\frac{N_t}{N} * (impurity - \frac{N_{t_R}}{N_t} * right\ impurity - \frac{N_{t_L}}{N_t} * left\ impurity)$ |
| CNN | Conv layer kernel size: 100, activation function: relu <br> Dense layer units: 6, activation function: softmax <br> optimizer: adam |
| Bi-LSTM | LSTM units: 64, activation function = tanh <br> Dense units: 4, activation function = softmax <br> optimizer: adam |

## 5 Conclusion

Resource barriers hindered many researchers from doing works on Bengali ABSA. We designed a benchmark dataset, BAN-ABSA, which can be used in further ABSA tasks and create more research scope. After several trial and error, we found out that BiLSTM works better than many other neural network models. We propose the BiLSTM architecture to be used in Aspect Based Sentiment Analysis. The results may not be too high compared to English ABSA. Modifications can be made in BiLSTM and add attention on attention to improve the overall performance. Researches in this field have just been started in the Bengali language. Numerous improvements can be made in this field.

## 6  Acknowledgement

We are very grateful to the SUST NLP Group and to the previous researchers who have worked in Bengali SA and ABSA. We are also very grateful to the researchers who have paved the way for NLP and Neural Networks.